\documentclass[11pt,a4paper]{article}
\usepackage{authblk}
\usepackage[hyperref]{acl2019}
\usepackage{times}
\usepackage{latexsym}
\usepackage{graphicx}
\usepackage{subcaption}
\usepackage{multirow}
\usepackage{amsmath}
\usepackage{xcolor}
\usepackage{colortbl}
\usepackage{verbatim}
\usepackage{enumitem}% http://ctan.org/pkg/enumitem
\newcommand{\resgatecol}[1]{\textcolor{orange!70!black}{#1}}
\newcommand{\ingatecol}[1]{\textcolor{green!40!black}{#1}}

\usepackage{url}

\aclfinalcopy % Uncomment this line for the final submission

\title{On the Realization of Compositionality in Neural Networks}

\author{Joris Baan$^1$, Jana Leible$^1$, Mitja Nikolaus$^2$, David Rau$^1$, Dennis Ulmer$^1$, \authorcr Tim Baumg{\"a}rtner$^1$, Dieuwke Hupkes$^{1,}$\thanks{\;\;Shared senior authorship}\; and Elia Bruni$^{3,*}$\\
\\
\texttt {\{joris.baan,jana.leible,david.rau,dennis.ulmer\}@student.uva.nl}\\
\texttt{mitja.nikolaus@posteo.de}\\
\texttt{baumgaertner.t@gmail.com}\\
\texttt{d.hupkes@uva.nl}\\
\texttt{elia.bruni@gmail.com}

} %\;\footnotemark[1]}

\affil[]{$^1$University of Amsterdam, $^2$University of T{\"u}bingen, $^3$Universitat Pompeu Fabra}
\date{}

\begin{document}
\maketitle
\begin{abstract}
We present a detailed comparison of two types of sequence to sequence models trained to conduct a compositional task.
The models are architecturally identical at inference time, but differ in the way that they are trained: our \textit{baseline} model is trained with a task-success signal only, while the other model receives additional supervision on its attention mechanism (Attentive Guidance), which has shown to be an effective method for encouraging more compositional solutions \citep{hupkes2018learning}.
We first confirm that the models with attentive guidance indeed infer more compositional solutions than the baseline, by training them on the lookup table task presented by \citet{livska2018memorize}.
We then do an in-depth analysis of the \textit{structural} differences between the two model types, focusing in particular on the organisation of the parameter space and the hidden layer activations and find noticeable differences in both these aspects.
Guided networks focus more on the components of the input rather than the sequence as a whole and develop small functional groups of neurons with specific purposes that use their gates more selectively. 
Results from parameter heat maps, component swapping and graph analysis also indicate that guided networks exhibit a more modular structure with a small number of specialized, strongly connected neurons.
\end{abstract}

\section{Introduction}
\label{introduction}
% Neural networks have proven to be extremely powerful models across a broad spectrum of domains \cite{lecun2015deep}. 
Sequence to sequence models (seq2seqs), a subset of neural networks that use sequences as input and output, have enjoyed great success in many NLP tasks such as machine translation \cite{bahdanau2014neural} and speech recognition \cite{graves2013speech}. Even though these feats indicate excellent generalization capabilities, the way seq2seqs generalize has found to be different from how humans do. In particular, seq2seqs lack of compositional understanding: the ability to construct new representations by combining familiar primitive components \citep[e.g.][]{szabo2012case}. Humans, instead, heavily rely on compositionality to learn complex functional structure efficiently \cite{schulz2016probing}. Once the primitive components are understood, a possibly infinite amount of novel combinations can be made, which allows for large scale generalization from a limited amount of examples \cite{fodor1975language}. For instance, sentences consist of words, which in turn consist of characters constructed from strokes.  
%A human face can be decomposed into two eyes, a 

%lack the skill of \textit{compositional} learning \cite{lake2017building}. Compositionality is the ability to construct new representations by combining familiar primitive components,  something that humans exploit for the efficient learning of complex functional
% . This hierarchical structure applies to almost everything and humans are known to exploit this to allow for the efficient learning of complex functional
%structure \cite{schulz2016probing}. 
%nose and a mouth. New faces can be made up or recognized by adjusting and recombining those elements.

%Vanilla seq2seq models with attention can produce

Recently, \citet{livska2018memorize} have shown how seq2seqs can produce many different fits on the training data using stochastic gradient descent, but rarely, if ever, find a compositional solution. The authors introduce a new data set called the \textbf{lookup table} task, which tests for out of distribution generalization. This data set will be discussed in more detail in Section~\ref{subsec:data}.

%This shortcoming in neural networks can be shown by testing on a data set that has a different distribution compared to the training set \cite{lake2017still}. 
%They accomplish this by using the same vocabulary but to construct the examples in such a way that the underlying components must be understood in order to obtain good performance. Such a data set will be discussed in more detail in Section \ref{related}.

As a remedy, \citet{hupkes2018learning} proposed Attentive Guidance (AG), a training technique which 
% Attentive guidance (AG) is a recently proposed technique by \citet{hupkes2018learning} that 
encourages seq2seqs to encode a more compositional solution without changing their internal architecture. AG provides additional information about the structure of the input sequence by supervising the attention mechanism of a model. As a result, the model is able to find what are the basic components of the lookup table task and how to combine them in a compositional manner. 
%which allows the model to discover the components that the sequence is constructed from.
%
%The intuition behind this is that the model is being provided with additional information about the structure of an input sequence. In other words, the components that a sequence is constructed from are explicitly uncovered using the attention mechanism.

%
Thanks to this work, we are now in the unique position of having a compositional (from now on \emph{AG}) and non-compositional (from now on \emph{baseline}) model that have identical architectures, but implement very different approaches to the same task.
In this paper, we compare those two models and aim to find structural differences between the way they organise their weights and form their representations, that could be indicative of compositional solutions.
In particular:
\begin{itemize}
    \item We show, through inspection of the parameter space and activations, that individual neurons in the AG model show a degree of specialization with respect to specific inputs that is unseen in baseline models. 
    \item We demonstrate, by substituting parts of both models with the corresponding component of its counterpart, which model sections contribute most to  the observed compositional behavior in AG models.
\end{itemize}

% the hidden layer activations, gate activations or organization of weights that reflect a model's ability of finding compositional solutions and distinguish it from a model that does not have this ability.
%find such compositional solutions. 
These differences confirm the findings of \citet{hupkes2018learning} that seq2seqs do not necessarily require big architectural adjustments to handle compositionality, since a network with identical architecture is capable of finding such a solution. 
Furthermore, these findings could be exploited to inform architectural changes in models, such that their priors to infer compositional solutions increase even when they are not provided explicit additional feedback on the compositional structure of the data.
% it would potentially allow for exploiting these differences to better induce compositionality in neural networks. 
%The question central in this paper is: what makes a model trained with AG exhibiting compositionality differ from a non-compositional baseline model?
%{\color{red} Sentence about potential importance to neuroscience?}\\

%e contributions of our paper can be summarized as follows:
%\begin{itemize}[noitemsep,topsep=0pt]

\section{Setup}\label{sec:setup}

In our experiments, we compare vanilla seq2seq with models that are trained with AG. 
Below, we briefly discuss both setups and the data we use for our experiments.

\subsection{Task}\label{subsec:data}
For our experiments, we use the lookup table composition task proposed by \citet{livska2018memorize}, which was created to test the compositional abilties of neural networks.
In this task, atomic lookup tables are created as to define a unique mapping from one binary string to another binary string of the same length. 
These atomic tables are then applied sequentially to a binary input string and yield a binary string. 
To give an example: if $t1$(001) = 110 and $t2$(110) = 001, then the function $(t1\circ t2)$(001) = 001 can be computed as a composition of $t1$ and $t2$. See Table \ref{tab:lookup} for a more comprehensive example.

Following \citet{hupkes2018learning}, we generate eight atomic lookup tables with strings of length 3 and use them to produce all 64 possible length two compositions. This forms the basis of the dataset that all experiments were performed on.
To test the model's ability of generalization on a more granular level, we compose four test sets with an increasing level of difficulty. For the first test set, we remove 2 out of 8 inputs for every composition (\textit{heldout inputs}). For the second and third testset, we remove 8 random table compositions from the training set (\textit{heldout compositions}), as well as all compositions that either contain $t7$ or $t8$ (\textit{heldout tables}). Finally, we create a test set by removing all compositions that contain a combination of tables $t7$ and $t8$ from the training set (\textit{new compositions}). 
The nature of the tasks requires the models to make use of the underlying compositionality. If this structure is not exploited, it is impossible to reliably find the correct solutions for the test data.
For more details, we refer to \citet{livska2018memorize} and \citet{hupkes2018learning}.
%In the original experiments, \citet{livska2018memorize} tested model performance only on \textit{heldout inputs}, whereas \citet{hupkes2018learning} created and tested on all four test sets.
\begin{table}[ht]
    \centering
\begin{tabular}{ c  c  c  }
    Atomic & Atomic & Composed \\
   $t1$ & $t2$ & $t1\circ t2$ \\ 
$000 \rightarrow 111 $ & $000 \rightarrow 100$ & $000 \rightarrow 011$\\
  
$001 \rightarrow 010 $ & $001 \rightarrow 101$ & $001 \rightarrow 110$\\

$010 \rightarrow 101 $ & $010 \rightarrow 110$ & $010 \rightarrow 100$\\
      \dots & \dots & \dots \\
\end{tabular}

    \caption{Example for atomic lookup tables ($t1$ and $t2$) of length 3 and a composition of length 2 ($t1\circ t2$).}
    \label{tab:lookup}
\end{table}

% More general title that doesn't collide with 3.2 and leaves room for more related literature

\subsection{Baseline}

The baseline model consists of an encoder-decoder architecture with an attention mechanism \cite{bahdanau2014neural} and Gated Recurrent Units (GRU)\footnote{We also trained models with Long-Short Term Memory units \cite{hochreiter1997long} but found the results to be very similar and therefore decided to omit the latter from this work.} \cite{cho2014learning}.

GRUs compute the hidden activations $h_t$ based on the previous hidden state $h_{t-1}$ and the representation of the current input $x_t$ in the following way (biases were omitted for clarity):
\begin{equation*}
    \begin{split}
    \ingatecol{z_t} &= \sigma(W_{iz} x_t + W_{hz} h_{t-1}) \\
    \resgatecol{r_t} &= \sigma(W_{ir} x_t + W_{hr} h_{t-1}) \\
    \tilde{h}_t &=\text{tanh}(W_{ih} x_t + W_{hh}(\resgatecol{r_t} \circ h_{t-1}))\\
    h_t &= (1 - \ingatecol{z_t}) \cdot h_{t-1} + \ingatecol{z_t} \cdot \tilde{h}_t,
    \end{split}
\end{equation*}

where we call $\ingatecol{z_t}$ and $\resgatecol{r_t}$ the activations of the \ingatecol{\emph{update gate}} and \resgatecol{\emph{reset gate}}, respectively.

\subsection{Attentive Guidance}

The AG model used in this work is identical to the baseline model in terms of architecture. 
The only difference occurs during the training procedure, where an additional loss term is enforced on the weights of the attention mechanism at decoding time step $t$ for input token $i$, $\hat{a}_{i, t}$:
\begin{equation*}
    \mathcal{L}_{\text{AG}} = \frac{1}{T}\Big(\sum_{t=1}^T\sum_{i=1}^N-a_{i, t}\log \hat{a}_{i, t} \Big),
\end{equation*}

where $\hat{a}_{i, t}$ denotes the target attention weights.
The attention loss is computed with an additional set of labels, that express how the input should be segmented and in which order it should be processed.
\citet{hupkes2018learning} show that providing  this additional supervision consistently improves the solutions found for the lookup table task: the guided models were found to have perfect generalization capabilities on the \textit{heldout compositions} and \textit{heldout inputs} and also perform well on \textit{heldout tables} and \textit{new compositions}. As inputs are supposed to be processed sequentially in our case, the target attention pattern is strictly monotonic, i.e. the target attention weights over the sequence are realized in a diagonal matrix.

\subsection{Experiments} \label{sec:experiments}
We train five baseline and AG models with the same hyperparameters and the Adam optimizer \cite{kingma2014adam}. Given the small vocabulary, we use an embedding size of 16 and a hidden size to $512$. All models were trained for a maximum of $100$ epochs with an attention mechanism, determining attention weights by using a multi-layer perceptron. Models were selected by their best accuracy on a held-out set. A comprehensive list of model performances on the different sets can be found in the Appendix.
%online\footnote{AnonymousURL
%See \url{https://github.com/Kaleidophon/machine-zoo}
The model implementations themselves stem from the \verb|i-machine-think| codebase.\footnote{Available under \url{https://github.com/i-machine-think/machine}.}

%\section{Experiments}

In the following, we perform three different suits of experiments. Firstly, we examine the parameter space of both models (Section \ref{sec:param-space}). Secondly, we take a closer look at the activations of single neurons and the GRU gates (Section \ref{sec:hidden-act}). Lastly, in Section \ref{sec:ablation}, we perform two different ablation studies: we make components of one model interacting with the components of the other and we distill the network via strongly connected neurons.

\section{Inspecting the Parameter Space}\label{sec:param-space}

\label{experiments}

In this section, we look at the parameter space of the baseline and AG models.
All discoveries regarding the parameter space 
were validated by comparing 5 runs of the same model class to make sure 
that observed differences can be ascribed to the differences in models and not 
different weight initializations.
%All experiments regarding the parameters were done separately for each layer of the networks, as well as within models and between model classes. This means that for example baseline model 1 would be compared to baseline model 2. This was done to validate that discoveries can actually be ascribed to AG vs baseline instead of stemming merely from a different initialization of weights or similarly irrelevant phenomena. 
%{\color{red} There is a difference in writing style: using vs avoiding 'we'. Should this be more uniform throught the paper?}

\subsection{Weight Inspection}
To gain a better understanding of the organization of the weights, we generated weight heat maps with the y-axis representing the weights going from all neurons to one neuron of the next layer (incoming weights) and the x-axis the weights going from one neuron to all neurons of the next layer (outgoing weights). Neural networks are known to be good at distributing their weights rather than have strong spatial organization, which makes it interesting to see whether such heat maps would reveal any differences in the organization of weights between AG and baseline models \footnote{Note that we do not normalize reported weights or activations by the activity of the 'pre-synaptic' neurons connected to it. This would be interesting to explore in future research, since a neuron's activation and the importance of its weight is in part dependant on the mean activation of its predecessors.}.

\begin{figure}[t]
     \centering
  \begin{minipage}[t]{\textwidth}
     \includegraphics[width=.45\textwidth]{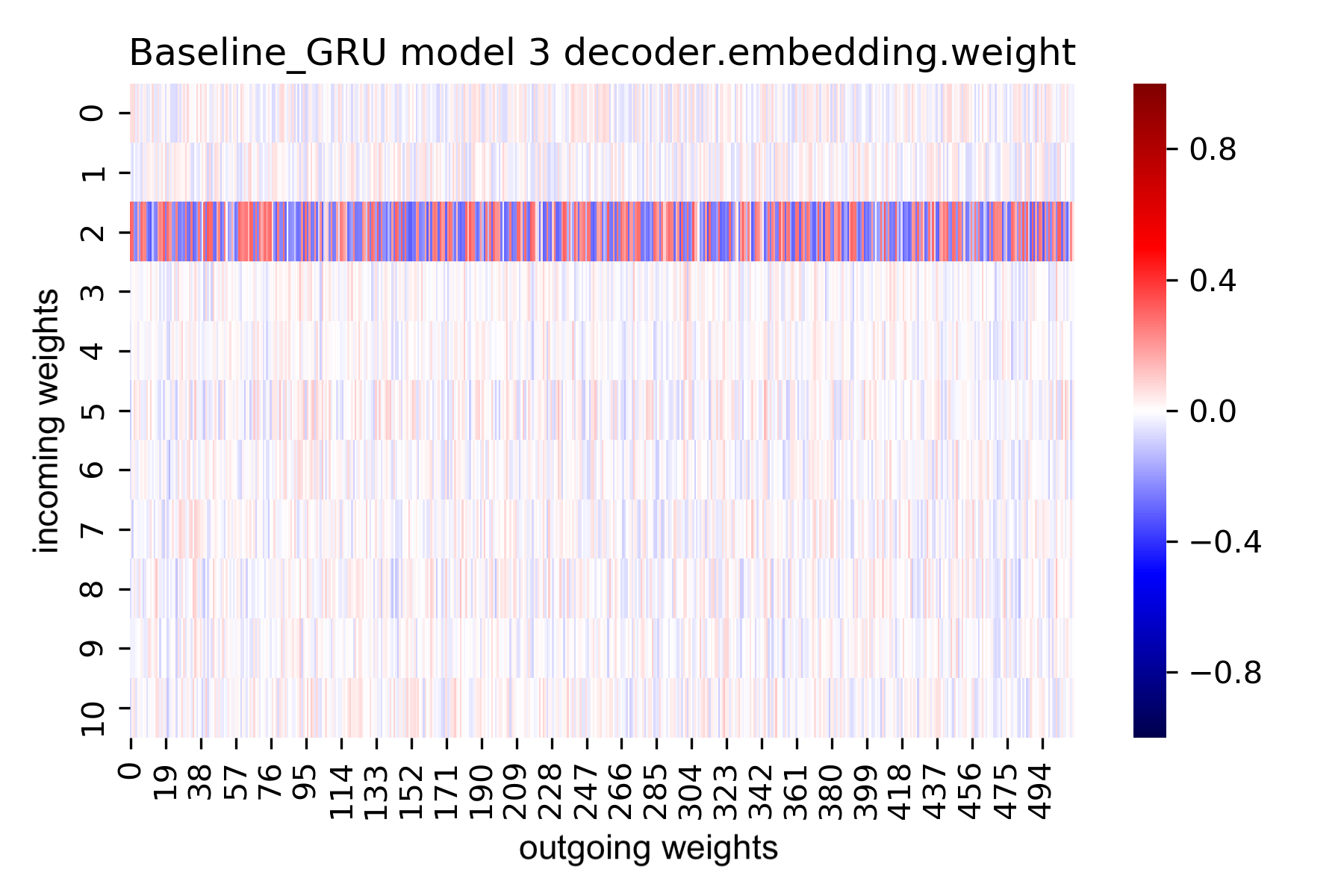}
  \end{minipage}
  
   \begin{minipage}[t]{\textwidth}
     \includegraphics[width=.45\textwidth]{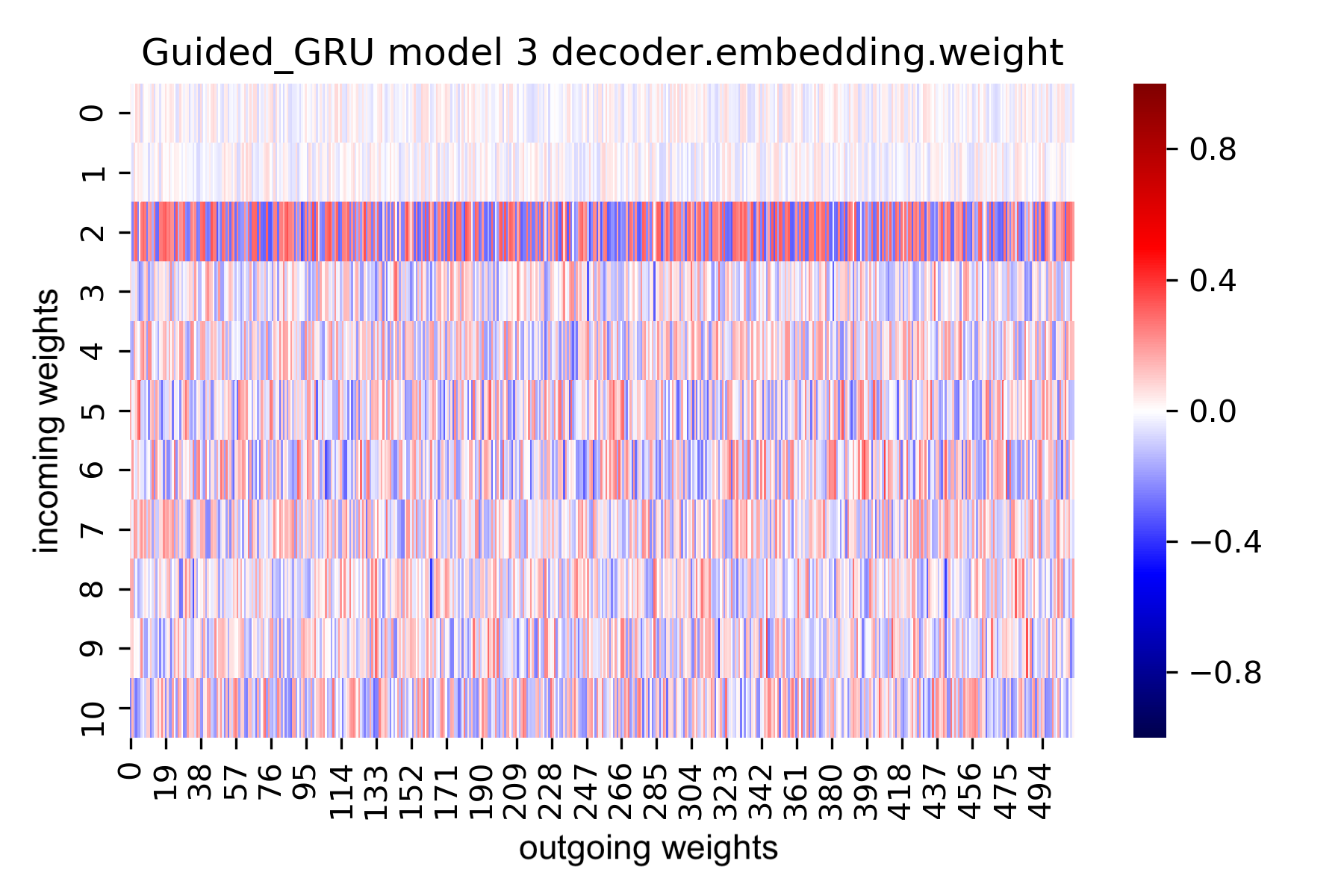}
  \end{minipage}
  \caption{Heatmap of the decoder embedding weight values. Outgoing weights correspond to weights going from the embedding to one decoder output neuron, and incoming weights to all weights going from the embedding to all decoder output neurons. (Best viewed in color)} 
\label{fig:heatmap}
\end{figure}

\begin{figure}
\centering
\begin{subfigure}[b]{0.49\textwidth}
\centering
\includegraphics[width=.49\columnwidth]{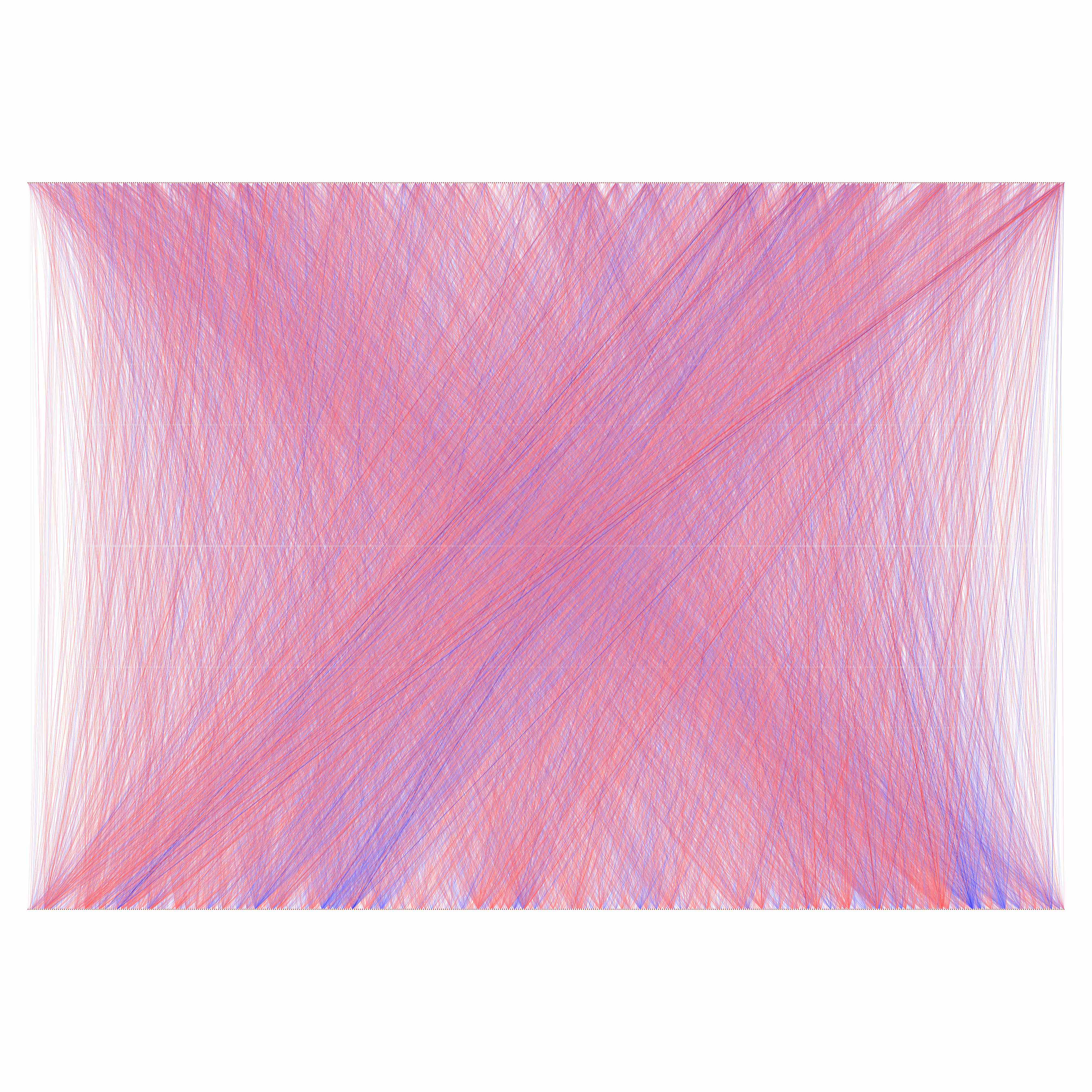}
\includegraphics[width=.49\columnwidth]{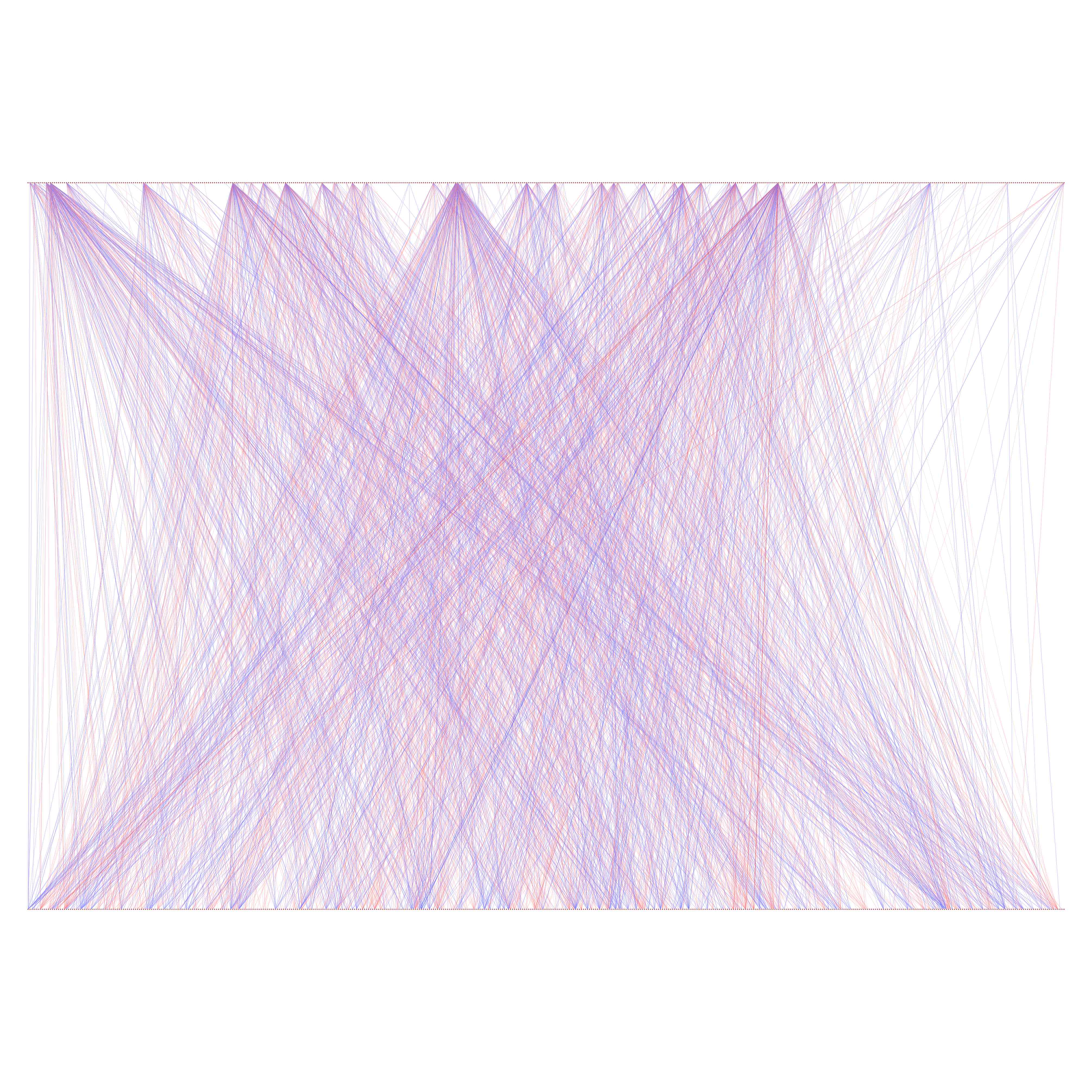}\\\vfill
\caption*{Left: baseline. Right: AG. Encoder update gates $W_{hz}$.}
\label{subfig:graph-encoder}
\includegraphics[width=.49\columnwidth]{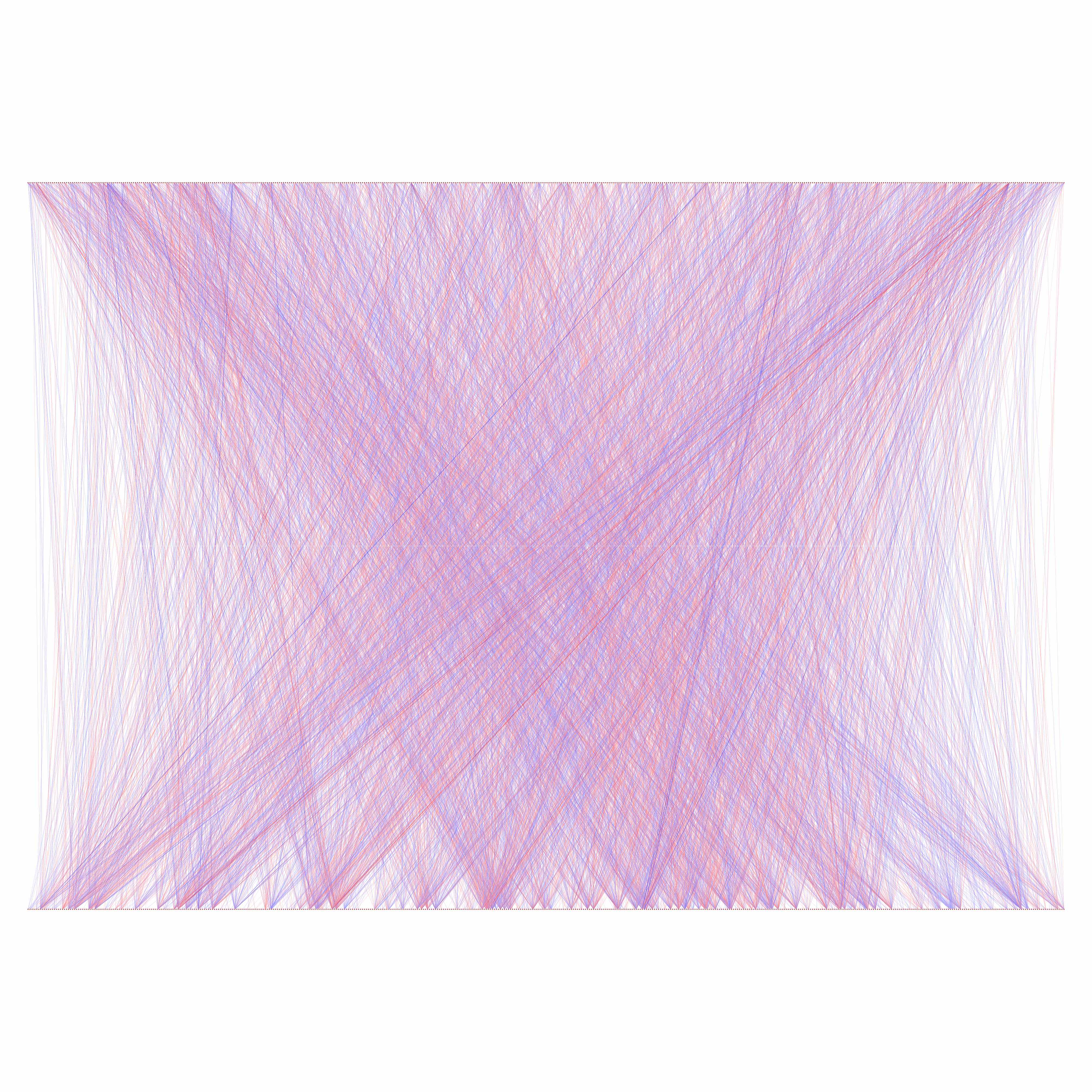}
\includegraphics[width=.49\columnwidth]{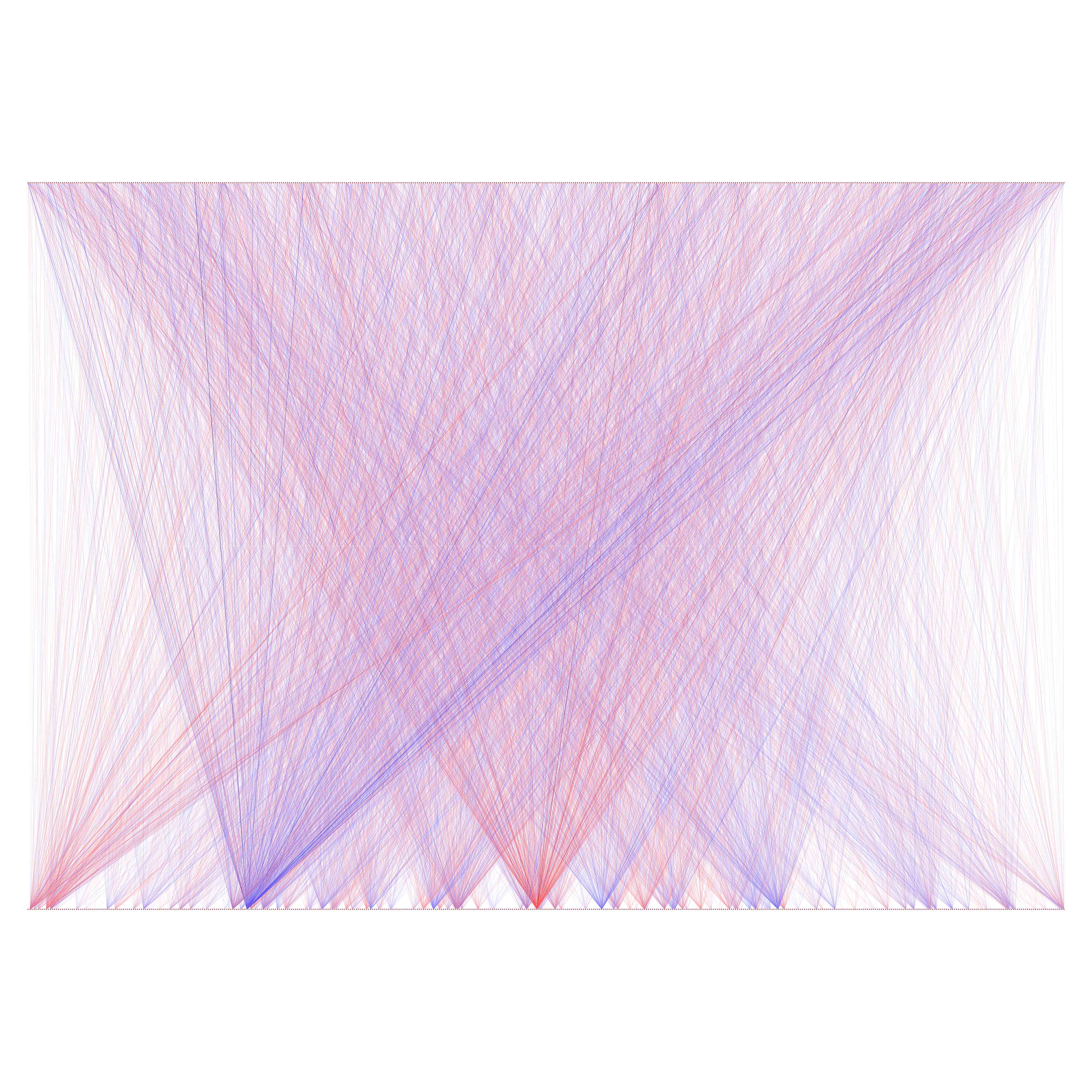}
\caption*{Left: baseline. Right: AG. Decoder update gates $W_{iz}$.}
\label{subfig:graph-decoder}
\end{subfigure}%
\caption{ Visualization of weight matrices $W_{hz}$ of the encoder and $W_{iz}$ of the decoder.  Weights going from the previous to the next layer are represented by lines going from bottom to the top. The color reflects the weight value, where blue denotes negative, red positive and white zero. (Best viewed in color)}
\label{fig:graph}
\end{figure}

The most striking difference between AG and baseline arises for the decoder embedding, as can be seen in Figure~\ref{fig:heatmap}. The baseline model exhibits small weights whereas the AG model shows bigger weights in rows 2-10. Row number two is an exception, since it is equally strong for both networks. This might be explained by the fact that this row represents the start-of-sequence (SOS) token, which could be sending a stronger error signal for both models. 

\subsection{Neural Connectivity}
\begin{figure}
\centering

\begin{subfigure}[b]{0.5\columnwidth}
\centering
\includegraphics[width=\columnwidth]{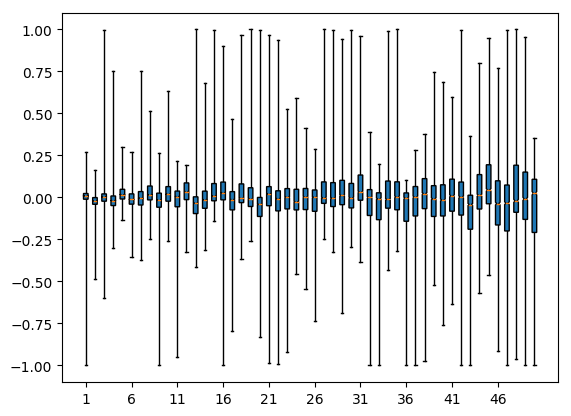}
\caption{Encoder: Baseline}
\label{subfig:activation-dists-gru-encoder}
\includegraphics[width=\columnwidth]{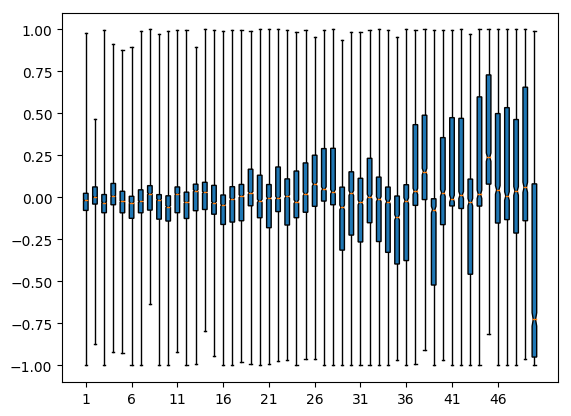}
\caption{Decoder: Baseline}
\label{subfig:activation-dists-gru-decoder}
\end{subfigure}%
\begin{subfigure}[b]{0.5\columnwidth}
\centering
\includegraphics[width=\columnwidth]{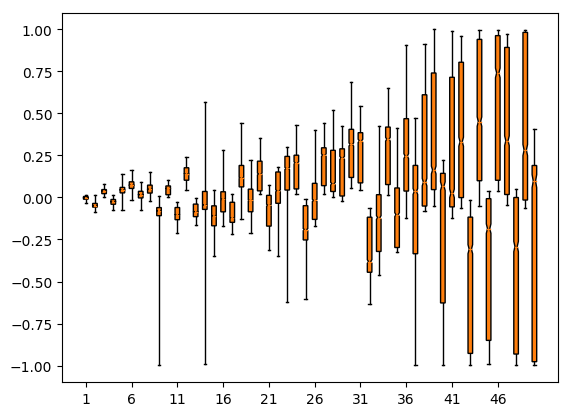}
\caption{Encoder: AG}
\label{subfig:activation-dists-gru-encoder-ag}
\includegraphics[width=\columnwidth]{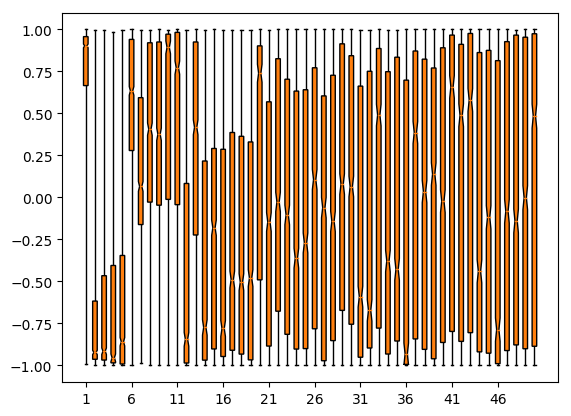}
\caption{Decoder: AG}
\label{subfig:activation-dists-gru-decoder-ag}
\end{subfigure}

\caption{Distributions of activation values for $50$ randomly sampled neurons for baseline (blue) and AG (orange) for both encoders (top) and decoders (bottom). Whiskers show the full range of the distribution. (Best viewed in color)}
\label{fig:activation-dists}
\end{figure}

Since the heat maps of the weight matrices for larger layers were hard to interpret, we explored a more intuitive visualization of the network's parameter space. We took neurons as nodes, and weights between neurons as edges. The thickness and color of an edge represents the magnitude of the weight. To prevent clutter, we applied thresholding to remove edges that corresponded to weak weights. For the encoder and decoder, we used a threshold of $\pm$0.2 and $\pm$0.17 respectively, which corresponds on average (between AG and vanilla models) to the strongest one percent of the weights.

The goal is to understand how the parameter space is structured and to see whether any differences between AG and baseline models can be found, for example, because of a stronger modularity, grouping or specialization of neurons in AG models. 
%An important note for the graph analysis is that, when neurons are close together in space, this does not necessarily indicate that these neurons are similar. Instead, they merely happen to be in a neighboring row in the weight matrix. % , which is perhaps caused by the order of tokens in the vocabulary. 

% When visualizing weights in an more intuitive way we encountered a pattern that seems to be specific for the compositional models.

Figure \ref{fig:graph} depicts the update gate weights $W_{hz}$ of the encoder on the top and the weights $W_{iz}$ of the decoder at the bottom. The weights of the previous layer to the next are represented by edges going from bottom to top. The most striking difference is that the baseline weights seem much more cluttered, whereas the AG model exhibits a few distinct, strongly polar neurons - neurons whose weights are on average negative or positive. 
%In the AG models decoder highly connected neurons occur at the bottom layer (input), whereas in the encoder these neurons are located in the top layer (weights to new hidden state). The same pattern holds for both, the $W_{hh}$ weight matrix.
Neurons that have many strong connections occur in the top layer of the encoder of the AG model in $W_{hz}$ and $W_{hr}$. Similarly, strong connected neurons can be found at the bottom layer of the AG's decoder in $W_{iz}$ and $W_{ir}$.
The finding of highly connected neurons seems to further reinforce the hypothesis that AG models learn to specialize using fewer but more strongly connected neurons, which could help to learn a more modular solution. %TODO EXPLAIN TOP vs BOTTOM IN ENCODER VS DECODER

Another interesting phenomenon that holds for both models can be observed by looking at the difference between update and reset gates of the same network (not shown here for the sake of space): The polarity of the neurons that are on average strongly positive or negative are inversely related. A possible explanation for this is that, when information from the current hidden state is to be retained, that same part is being reset in the previous hidden state.

\section{Analyzing Activations}\label{sec:hidden-act}

While analyzing the model weights gives us insight into the general trained structure of the model, analyzing activations lets us examine how the different model types respond to certain inputs. We thus try to identify groups of neurons that specialize to respond to certain inputs and provide further inside into the GRU's gating behavior.

\subsection{Functional Groups} \label{sec:diagnostic_classifiers_functional_groups}
We hypothesize that solving the task compositionally is done by distinct groups of neurons in the network. Each group addresses different functionalities. For example, a group of units in the encoder could be responsible for representing the presence of the current table in the input sequence, as proposed in the previous section.

An indicator for this behavior can be seen in Figure \ref{fig:activation-dists}, where we sampled 50 random neurons from 
the encoder and decoder of both models and tracked their activation values emitted over the samples in the test set.
We can see that in contrast to the baseline, some neurons of the AG only produce activations in specific value ranges, which could be a hint for a potential specialization. The same can be found inside the AG's decoder, although most of the neurons sampled seem to cover the whole value range during processing.

To test this hypothesis, we analyze which hidden activations are crucial for correctly predicting the current table at a time step. The baseline model is expected to not be able to predict the presence of single tables because it fails to see the tables as parts of a compositional task and instead memorizes the combinations it has encountered during training.

% \footnote{The SAGA solver \cite{defazio2014saga} implementation of \textit{scikit-learn} was used for the classification task.}

In a first experiment, we use diagnostic classification \citep[DC,][]{hupkes2018visualisation}, which consists in training linear classifiers on the hidden activations to predict a certain feature. 
%(\citet{hupkes2018visualisation}, \citet{dalvi2019one}, \citet{conneau2018you}),
 In this case, we use the encoder's activations to predict the table in the input sequence of the corresponding time step. For example, if the input was `000 t1 t2', we trained the classifier to predict `t1' for the encoder activations of the second time step and to predict `t2' for the activations of the third time step. 
% which we call \emph{Diagnostic Classifiers} (DCs). 
 Similarly to the methodology of \citet{dalvi2019one}, we subsequently added units to a set, depending on the absolute weight they were assigned in the diagnostic classifier.\footnote{However, unlike \citet{dalvi2019one}, we do not use any regularization on the DC to contrast the different degrees to which information is distributed across neurons in the two model types.} After each addition, we re-calculated the accuracy for the prediction. This process was repeated until 95~\% of the overall accuracy (with all units) is reached, the resulting subset of units forms the \emph{functional group}.\footnote{Applying the methods development by \citet{lundberg2017unified} seems to confirm the responsible neurons we found, but selects more neurons and gives less consistent results, which we trace back to the extensive approximations required and some model assumptions (e.g.\ feature independence) being violated.}

The results are shown in the first row of Table \ref{tab:predicting-tables-gru}. All numbers are averaged over the five trained models.
% It is possible to predict the current table with both the guided and the baseline models. 
Some differences arise in the functional group size of the models: While for the baseline models on average 35 units are required to make a good prediction, the information is stored in only 2 units in the guided models. 
% We clearly detect a functional group of units in the encoder that represents the current input table.

To verify whether the units in the functional group are actually important units in the model, we further analyzed the strengths of the weights connected to each of the units. On average, 93\% of the units in the functional group of the AG models can be found in the top 5\% of the units with the strongest absolute weight values. We conclude that the units of the functional group are highly connected and thus very likely to play an essential role in the functionality of the model.

\begin{table}[ht]
\begin{center}
\begin{tabular}{@{}rlll@{}}
\hline
In & Model & Accuracy & \#Units \\
\hline
\multirow{2}{*}{$h_t^{\text{enc}}$} & BL  & .93 (.98) & 35\\
 & AG & .98 (1.) & 2 \\
\hline
\multirow{2}{*}{$\ingatecol{z_t^{\text{dec}}}$} & BL & .51 (.53) & 52\\
 & AG & .96 (1.) & 22.2 \\
\hline
\multirow{2}{*}{$\resgatecol{r_t^{\text{dec}}}$} & BL & .50 (.52) & 44  \\
& AG & .96 (1.) & 20.8\\
\hline
\end{tabular}
\caption{Performance of diagnostic classifiers for predicting the current input table with the hidden activations ($h_t^{\text{enc}}$)  of the encoder, the input gate activations ($\ingatecol{z_t^{\text{dec}}}$) or the reset gate activations ($\resgatecol{r_t^{\text{dec}}}$) of the decoder of the baseline (BL) and Attentive Guidance (AG) model. The third column shows the accuracy when predicting using the functional group of units and in brackets the accuracy when using all units. The fourth column displays the average number of units in the functional group across different runs (which can be either hidden units or gate activations).}
\label{tab:predicting-tables-gru}
\end{center}
\end{table}

Assuming that the information of the current table being stored in the encoder activations is used by the decoder to perform according calculations, we expect that by using the gate activations of the decoder it is also possible to predict the current input table. We use the same methodology as in the previous experiment, with the only difference that the inputs for the diagnostic classifier are the activations of the decoder gates. Results are shown in the second and third rows of Table \ref{tab:predicting-tables-gru}. Using all gate activations of the update or the reset gate of GRUs, we are able to perfectly predict the current table in the guided models. With the baseline model, an accuracy of only around50~\% is reached.\footnote{Accuracy with a majority classifier for the task is 12.5~\%.} The size of the functional groups in the guided models is remarkably larger than with the encoder hidden activations, showing that the information is more distributed over the gates.
This difference can be explained by the fact that the gates are not mainly representing information, but using represented information to perform calculations. Further, distribution of information across the gates is more likely because a gate activation affects only one hidden unit while a hidden layer activation can possibly affect all gates in the upcoming time step \cite{hupkeszuidema2017}. 

In another experiment, we aim to predict the current time step with the activations of the encoder.\footnote{For example, if the input was `000 t1 t2', we trained the classifier to predict `0' for the encoder activations of the first time step, `1' for the encoder activations of the second time step and `2' for the activations of the third time step.} We assume that counting is an essential part of solving the task in a compositional manner. The methodology is the same as in the previously described experiments. The result pattern (cf. Table \ref{tab:predicting-timesteps}) can be compared to the first experiment: Using all units, it is possible to predict the time step with all models, but, in the guided attention models, the information is more concentrated in functional groups than units.
%, here consisting of only two units.

\begin{table}[ht]
\begin{center}
\begin{tabular}{@{}rlll@{}}
\hline
In & Model & Accuracy & \#Units \\
\hline
\multirow{2}{*}{$h_t^{\text{enc}}$} & BL & .95 (.98) & 40  \\
& AG & 1.0 (1.0) & 2 \\
\hline
\end{tabular}
\caption{Performance of diagnostic classifiers for predicting the current time step with the hidden activations ($h_t$) of the encoder. 
The second column shows the accuracy when predicting using the functional group of units and in brackets the accuracy when using all units. The third column displays the number of units in the functional group.}
\label{tab:predicting-timesteps}
\end{center}
\end{table}

These results, implying that some neurons specialize in tracking the current time step and reacting to distinct inputs, demonstrate that the AG model uses information about the current table in the decoder to perform operations in a compositional way (treating the tables as distinct parts). The baseline model does not show distinct activation patterns in the gates for specific tables.

\subsection{Gating behavior}
Based on the findings in previous section, we also expect the usage of reset and update gate to be significantly different for the two models under scrutiny.
To study this, we use a technique introduced by \citet{Karpathy2015}, that considers, for each gate in the network, the fraction of samples for which it is left-saturated (activation smaller than 0.1) or right-saturated (activation greater than 0.9), where being left-saturated corresponds to being closed and right-saturated to being open.

\begin{figure*}
\centering

\begin{subfigure}[b]{0.49\textwidth}
\centering
\includegraphics[width=\columnwidth]{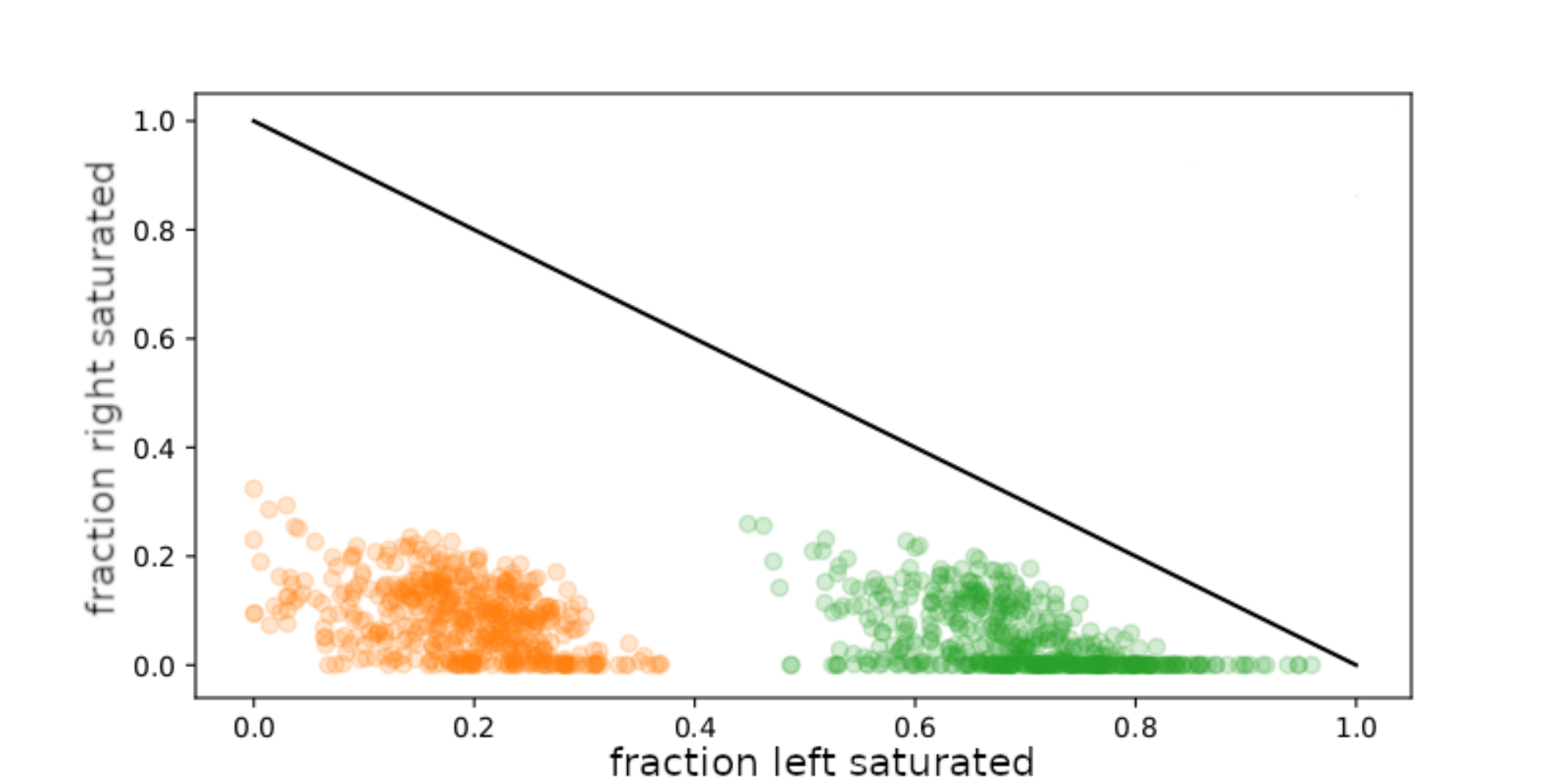}
\caption{Encoder: Baseline}
\label{subfig:gate-saturation-gru-encoder}
\includegraphics[width=\columnwidth]{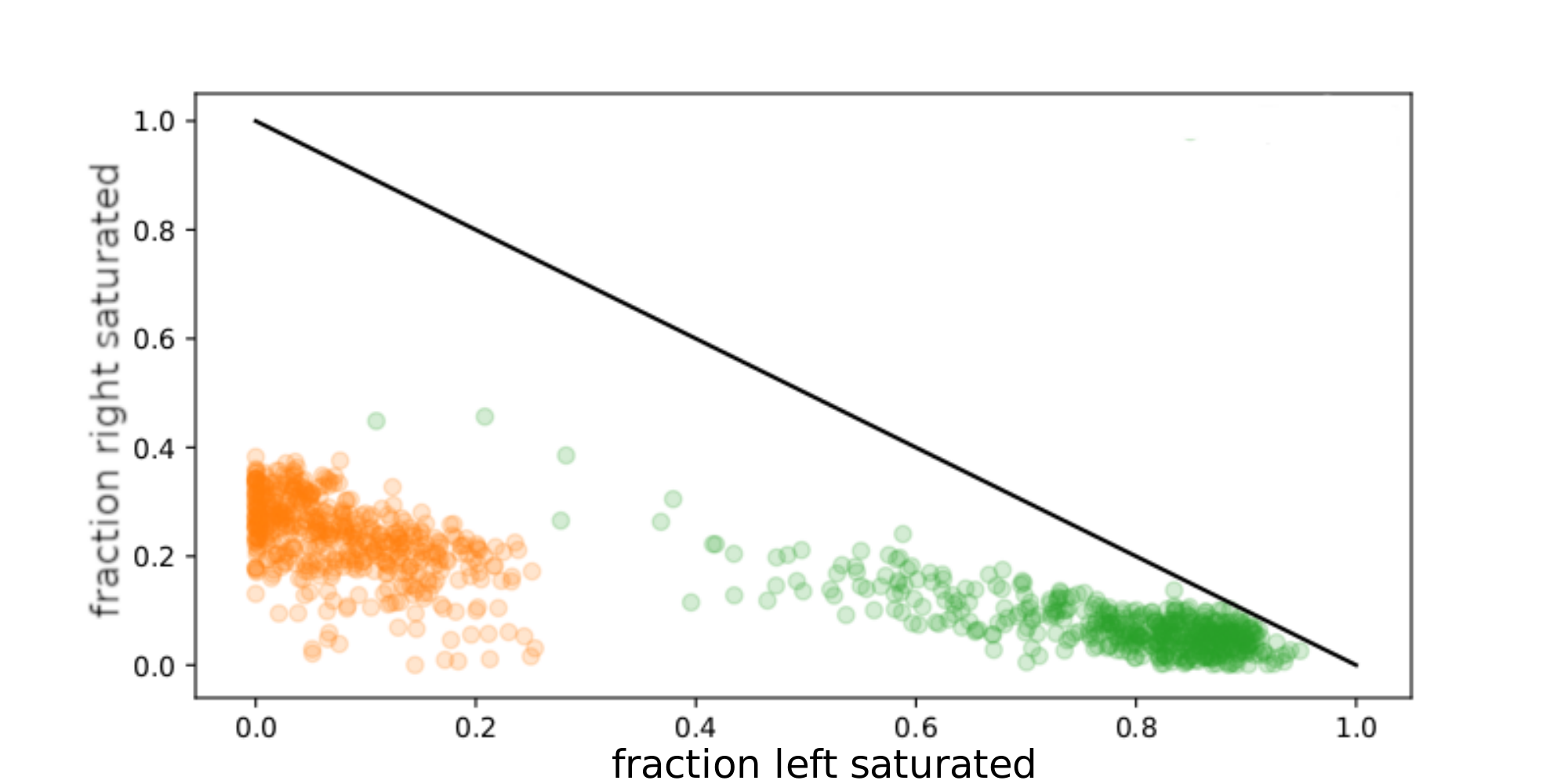}
\caption{Decoder: Baseline}
\label{subfig:gate-saturation-gru-decoder}
\end{subfigure}%
\begin{subfigure}[b]{0.49\textwidth}
\centering
\includegraphics[width=\columnwidth]{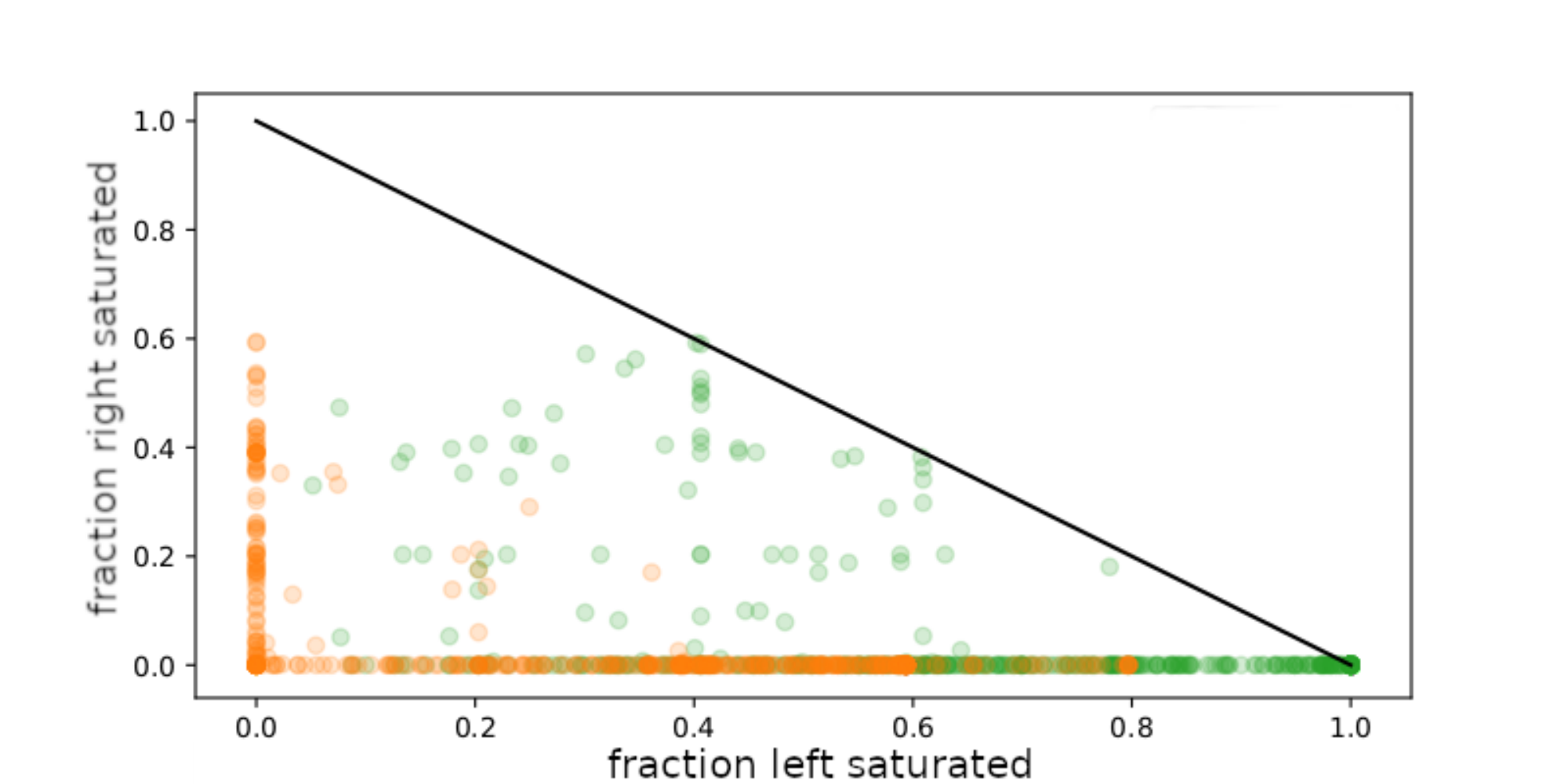}
\caption{Encoder: AG}
\label{subfig:gate-saturation-gru-encoder-ag}
\includegraphics[width=\columnwidth]{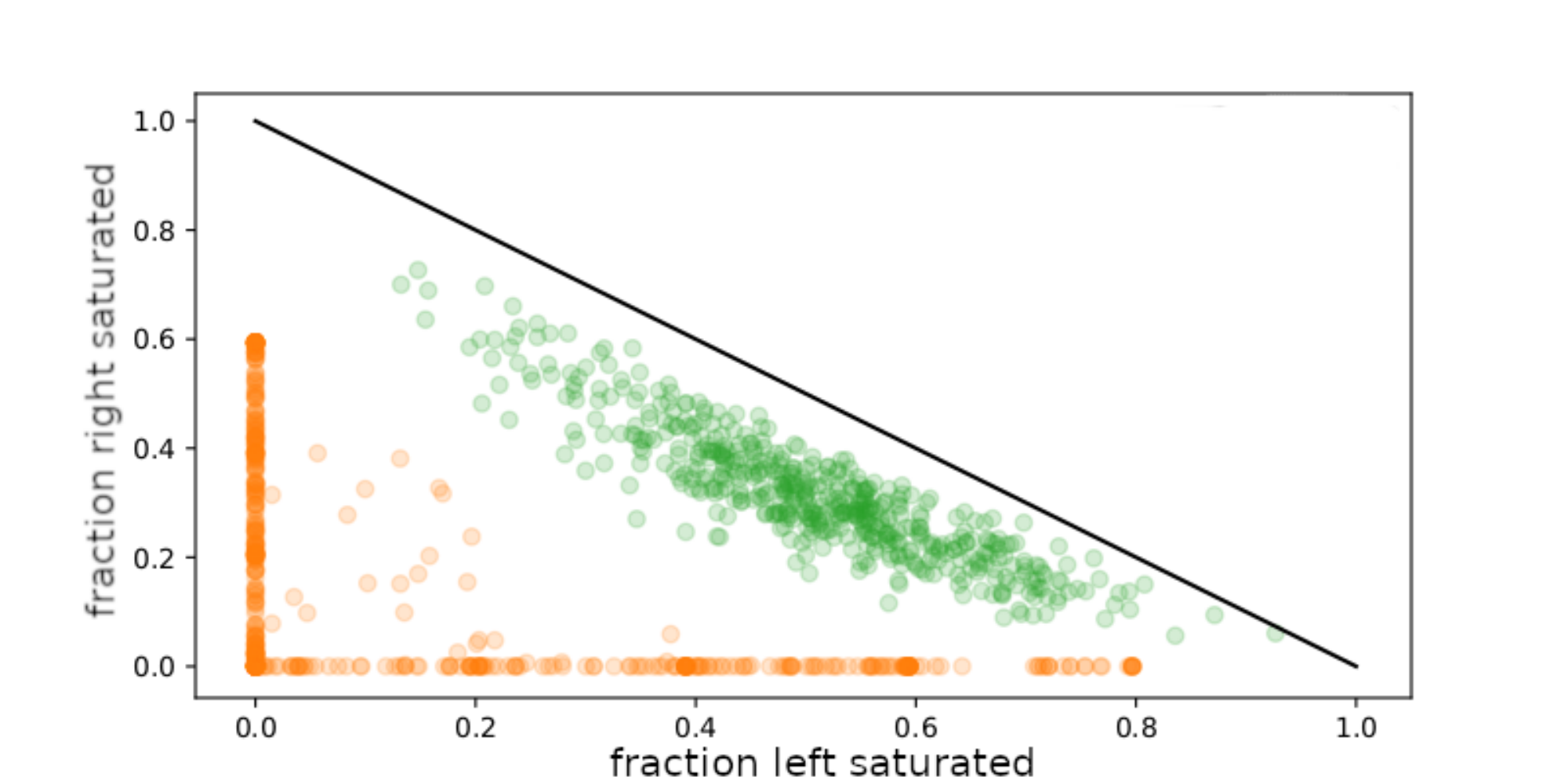}
\caption{Decoder: AG}
\label{subfig:gate-saturation-gru-decoder-ag}
\end{subfigure}

\caption{Gate activation plots for \resgatecol{reset gate $r_t$} and \ingatecol{update gate $z_t$}. (Best viewed in color)}
\label{fig:gate-saturations}
\end{figure*}

We show the results in Figure~\ref{fig:gate-saturations}.
The plots reveal a clear difference between the usage of gates in the baseline and the guided models.
The guided models seem to be more distinct in their gate activation: Activations tend to stick to the axes. Values close to the diagonal mean that the respective gate is mostly saturated, values close to the axes reveal gates that are either saturated to one side or not saturated.
Values in between, mostly seen in the baseline models, indicate gates that are rarely saturated.

The activation pattern of the \ingatecol{update gates} shows clear differences between the baseline and the AG models. In the baseline, they are mostly left saturated, which means that new information is rarely incorporated in the calculations, in both the encoder and decoder.
In the encoder of the AG model, most gates are also only left-saturated, but there is a considerable amount of outliers which could be some units that are highly specialized to specific inputs. 
In the AG decoder, some gates are also often right-saturated, allowing for the intake of new information. One possible interpretation is that the gates in the AG decoder model selectively allow the relevant input for the current time step to be included in the calculations.

\section{Ablation studies}\label{sec:ablation}
In this section, we first swap components of both models and measure the effects on the models' performances to identify crucial model parts. We then check performance of the AG model when only its strongly connected neurons are used.

\subsection{Component substitution}\label{sec:substitution}
To understand to what extent specific components of a seq2seqs contribute to compositionality, we take components from one trained model and place them into the other model. We freeze the weights of the replaced component to prevent any re-learning, and retrain the resulting model using its original training procedure. We extract a total of eight different components. The entire encoder and decoder as well as embeddings, internal GRU weights for input to hidden ($W_{ih}$), and recurrent hidden to hidden ($W_{hh}$) for both encoder and decoder. The experiment is twofold: components taken from a model trained with AG are placed into a baseline model and retrained without AG to examine whether a baseline can still learn a compositional solution. Additionally, components from a baseline model are placed into an AG model which is retrained with AG to examine whether the model can still learn a compositional solution without being able to adjust the parameters of the baseline component. We tune 16 models for each new component by retraining the remaining original parts for a maximum of 100 epochs, with a learning rate of 0.001 to allow for limited adjustment.

When retraining an AG model with a frozen baseline component, we expect the performance to drop when that component is important for a compositional solution, as the network is apparently unable to recover itself. Conversely, if a baseline model with a frozen component extracted from an AG model is retrained without AG and performs better, that component might contain weights organized in such a way that it forces the baseline model to retrain itself in a more compositional manner.
Table \ref{table:swap} shows the results of substituting components on \textit{new compositions}, which is considered the hardest compositional task. None of the baseline models given a frozen component from an AG model are able to retrain themselves such that they significantly increase performance. Thus, we left those results out of Table \ref{table:swap} for the sake of brevity. 

For the AG models with frozen baseline components, the encoder embeddings seem irrelevant for a compositional solution: using baseline embeddings result in a similar score. The decoder embeddings, however, do seem to play a role, as indicated by a much lower score than the original AG model. This seems to be in line with the differences in heat maps shown earlier in Figure~\ref{fig:heatmap}. Replacing the entire encoder results in a 80\% drop in accuracy to 0.167. Interestingly, the encoder hidden to hidden ($W_{hh}$) weight can be replaced without as big a drop, and using the baseline input to hidden ($W_{ih}$) weights actually improves the accuracy. Finally, replacing the decoder's $W_{ih}$ weights drops the accuracy to around 0.6, but doing the same for the decoder's $W_{hh}$ weights again results in an unexpected increase to almost 0.9. This seems to indicate that the $W_{ih}$ weights of the decoder play an important role in a compositional fit, as the model is unable to recover itself when using baseline decoder $W_{ih}$ weights. The increase in performance after replacing either the encoder's $W_{ih}$ or the decoder's $W_{hh}$ implies that training with AG actually produces suboptimal weights for these components. Perhaps the use of a frozen baseline component in a model retrained with AG acts as some kind of regularization and incentivizes the remaining components of the model to become more compositional. Another explanation could be that the AG loss does not provide an appropriate signal for all components, and should thus not be backpropagated to all of them. %Using a frozen component trained without AG could be considered as not backpropagating the AG loss to it.
\begin{table}[ht]
\begin{center}
\begin{tabular}{@{}rll@{}}
\hline
Model & Component & Accuracy (NC)  \\
\hline
AG          & -                 & .82 $\pm$ .12 \\
AG          & Encoder           & .17 $\pm$ .11 \\
AG          & Encoder Emb       & .75 $\pm$ .09 \\
AG          & Encoder $W_{ih}$  & .89 $\pm$ .05 \\
AG          & Encoder $W_{hh}$  & .79 $\pm$ .12 \\
AG          & Decoder           & .12 $\pm$ .05 \\
AG          & Decoder Emb       & .31 $\pm$ .07 \\
AG          & Decoder $W_{ih}$  & .60 $\pm$ .08 \\
AG          & Decoder $W_{hh}$  & .91 $\pm$ .03 \\
\hline
BL          & -                 & .01 $\pm$ .02 \\
BL          & Encoder           & .02 $\pm$ .02 \\
% Baseline    & Encoder Emb  & \textbf{0.031}  $\pm$ 0.026 \\
BL          & Decoder           & .02  $\pm$ .02 \\
% Baseline    & Decoder Emb  & \textbf{0.021}  $\pm$ 0.029 \\

\hline
\end{tabular}
\caption{Sequence accuracy on new compositions (NC). Accuracy is averaged over three models and depicted with its standard deviation. The model being retrained is specified the first column, the component taken from the opposite model and frozen specified in the second column.}
\label{table:swap}
\end{center}
\end{table}

\subsection{Neuron pruning}\label{sec:distill}
After showing in Section~\ref{sec:diagnostic_classifiers_functional_groups} that a few strongly connected neurons organized in functional groups carry out specific functions, we want to exhaust this observation and see if the model can still successfully solve the task by using \textbf{only} strongly connected neurons. We remove all weakly connected neurons, keeping only 5\% of neurons with the biggest weights of the encoder and decoder of the trained AG models respectively. Distilling the network in this way results in a performance drop to 12.4\% sequence accuracy on the new compositions task, averaged over all models. Re-training the network for 20 epochs fully restores the functionality and even yields better performance of 92.5\% on average compared to the full network with 82.3\%.  We retrained the model using the same parameters as in the main training procedure (see Section~\ref{sec:experiments}). 

The loss in performance that occurs when neurons are removed indicates that some functionality is distributed among weakly connected neurons. However, the fact that their functionality can be taken over by other neurons shows that weakly connected neurons do not play a crucial role. 

We conclude that most of the neurons do not contribute to a compositional solution at all and therefore only an extremely small subset of all neurons of the AG model suffices to solve the task after retraining. Those neurons exhibit strong weights and are specialized in functional groups. Networks that find a compositional solution seem to rather form a small number of highly specialized neurons than distributing functionality over the whole network.

\section{Conclusion}
\label{conclusion}

%{\color{red}Paragraph about potential importance to neuroscience/brain?}
%A seq2seq model with attention is not capable of compositional generalization in the lookup table task if trained in the standard way. 

Thanks to Attentive Guidance, seq2seqs are able to generalize compositionally on the lookup table task when, without it, they cannot \cite{hupkes2018learning, livska2018memorize}.
In this paper, we presented an in-depth analysis of the differences between an attention-based sequence to sequence model trained with and without Attentive Guidance. Any identified differences can contribute to our understanding of what makes seq2seq better compositional learners and help with the design of a new generation of compositional learning architectures.
%
%We analyzed both the hidden layer activations and the parameter space of the models and highlighted clear differences between the two models. 

Our main finding is that guided networks have a more modular structure: small subsets of well-connected neurons are responsible for specific functions. 
Having specialized neurons could be crucial to a compositional solution. 
We have also shown via component substitutions how these neurons 
seem to play a more crucial part in specific model components like the encoder / decoder gates and decoder embeddings, while playing a negligible role in others.

%{\color{blue} We need one more sentence about our findings.}
% The recurrent GRU weights are responsible for learning the weights for the update and reset gates for the hidden states, which regulate how memory (information from previous time steps) is incorporated into the current hidden state. We argue that, for a compositional solution, information from previous time steps in the encoder is more selectively let through because of more specialized neurons. Therefore, the recurrency of the encoder might be crucial to perform a compositional task. 

Future research could focus on exploiting the findings about modularity and specialization of neurons to investigate whether similar compositional solutions can be achieved without the explicit use of Attentive Guidance, such as recently shown by \citet{korrel2019transcoding}
Additionally, it would be interesting to find out why models with fewer parameters cannot learn to solve the lookup table task \citep{hupkes2018learning}, while we know from our distillation experiments that only 26 neurons in the encoder and decoder are needed to implement a perfect solution.

\section*{Acknowledgements}
DH is funded by the Netherlands Organization for Scientific Research (NWO),
through a Gravitation Grant 024.001.006 to the Language in Interaction Consortium. EB is funded by the European Union's Horizon 2020 research and innovation program under the Marie Sklodowska-Curie grant agreement No 790369 (MAGIC).

\bibliography{project_refs}
\bibliographystyle{acl_natbib}

\clearpage

\newpage
\appendix
\section{Model performances}
Sequence accuracy for the increasingly difficult tasks heldout compositions (HC), heldout inputs (HI), heldout tables (HT) and new compositions (NC) of the baseline (Tab. \ref{tab:model_performances_baseline}) and the AG (Tab. \ref{tab:model_performances_guided}) models.
\begin{table}[ht]
\centering
\begin{tabular}{rllll}
\hline
Run	& HC &	HI &	HT &	NC \\
\hline
1&	.25 &	.20 &	.04 &	.00 \\
2&	.16 &	.20 &	.06 &	.00 \\
3&	.25 &	.23 &	.04 &	.03 \\
4&	.22 &	.23 &	.03 &	.03 \\
5&	.23 &	.20 &	.07 &	.06 \\
\hline
\end{tabular}
\caption{\textbf{Baseline} GRU models}
\label{tab:model_performances_baseline}
\end{table}

\begin{table}[ht]
\centering
\begin{tabular}{rllll}
\hline
Run	& HC &	HI &	HT &	NC \\
\hline
1 &	1.0 &	1.0 &	0.85  &	0.69 \\
2 &	1.0 &	1.0 &	0.98  &	0.97 \\
3 &	1.0 &	1.0 &	0.88  &	0.81 \\
4 &	1.0 &	1.0 &	0.98  &	0.97 \\ 
5 &	1.0 &	1.0 &	0.94  &	0.91 \\
\hline
\end{tabular}
\caption{\textbf{Guided Attention} GRU models}
\label{tab:model_performances_guided}
\end{table}

\end{document}